\documentclass[runningheads]{llncs}
\usepackage[T1]{fontenc}
\usepackage{multirow}
\usepackage{booktabs}
\usepackage{colortbl}
\usepackage{makecell}
\usepackage{xcolor} 
\usepackage{bm}
\usepackage{tikz}
\usepackage{hyperref}

\hypersetup{
    colorlinks=true, 
    urlcolor=magenta,   
}
\usepackage{amsmath}
\usepackage{amsfonts}
\usepackage{graphicx,verbatim}
\usepackage{amssymb}
\usepackage{array}
\usepackage{color}
\usepackage{bbding}
\begin{document}
\title{LLM-driven Knowledge Enhancement for Multimodal Cancer Survival Prediction}

\author{
Chenyu Zhao\inst{1}$^{\star}$ \and
Yingxue Xu\inst{1}$^{\star}$ \and
Fengtao Zhou\inst{1} \and
Yihui Wang\inst{1} \and
Hao Chen\inst{1,2,3,4}\Envelope
}
\institute{Department of Computer Science and Engineering, The Hong Kong University of Science and Technology, Hong Kong SAR, China \and Department of Chemical and Biological Engineering, The Hong Kong University of
Science and Technology, Hong Kong SAR, China \and Division of Life Science, The Hong Kong University of Science and Technology,
Hong Kong SAR, China \and HKUST Shenzhen-Hong Kong Collaborative Innovation Research Institute, The
Hong Kong University of Science and Technology, Futian, Shenzhen, China \and State Key Laboratory of Nervous System Disorders, The Hong Kong University of
Science and Technology, Hong Kong SAR, China\\ 
    \email{cyzhao@connect.ust.hk, jhc@cse.ust.hk}}

\maketitle              
\begin{abstract}
Current multimodal survival prediction methods typically rely on pathology images (WSIs) and genomic data, both of which are high-dimensional and redundant, making it difficult to extract discriminative features from them and align different modalities.
Moreover, using a simple survival follow-up label is insufficient to supervise such a complex task. 
To address these challenges, we propose KEMM, an LLM-driven \textbf{K}nowledge-\textbf{E}nhanced \textbf{M}ultimodal \textbf{M}odel for cancer survival prediction,
which integrates expert reports and prognostic background knowledge. 
1) Expert reports, provided by pathologists on a case-by-case basis and refined by large language model (LLM), offer succinct and clinically focused diagnostic statements. This information may typically suggest different survival outcomes.
2) Prognostic background knowledge (PBK), generated concisely by LLM, provides valuable prognostic background knowledge on different cancer types, which also enhances survival prediction.
To leverage these knowledge, we introduce the knowledge-enhanced cross-modal (KECM) attention module. KECM can effectively guide the network to focus on discriminative and survival-relevant features from highly redundant modalities. 
Extensive experiments on five datasets demonstrate that KEMM achieves state-of-the-art performance. The code will be released upon acceptance.


\end{abstract}
\section{Introduction}
\label{intro}
Survival prediction is a complex ordinal regression task that aims to estimate the death risk of patients \cite{mlsasurvey,porpoise}, 
which typically benefits from incorporating pathological whole slide images (WSIs) and genomic data \cite{mcat,motcat,pibd} that complement each other. 
WSIs provide tissue and cell morphological information with high-resolution \cite{van2021deep,madabhushi2016image}.
For example, visual characteristics of the tumor microenvironment have been shown to correlate significantly with survival analysis, including cellular components such as fibroblasts and various immune cells, which can influence cancer cell behavior \cite{xiao2021tumor,chen2021clinical,yang2023cancer}.
Genomic data is also high-dimensional, offering quantitative molecular information that can indicate different molecular subtypes of tumors in prognostic analysis \cite{Zang2016,mcat,motcat}. The key to survival prediction is effectively utilizing survival-relevant information from each modality. 

Recent fruitful studies integrate WSI and genomic data to tackle the challenging survival prediction task. Nevertheless, two critical issues remain. First, both pathology and genomics modalities are highly redundant, significantly hindering multimodal alignment and integration for survival prediction. For example, the areas related to tumor microenvironment, crucial for survival assessment, occupy only a small portion of gigapixel WSIs ($100,000 \times 100,000$ pixels), while humans possess over 20k genes, with only a few strongly correlated with prognostic outcomes. Although previous works, \textit{e.g.}, MCAT \cite{mcat}, MOTCat \cite{motcat}, CMTA \cite{cmta} and SurvPath \cite{survpath}, attempt to capture cross-modal interactions between these two modalities, they still struggle to align interactions across modalities from the flood of information due to the nature of high redundancy. Second, although prototype-based models, e.g., PIBD \cite{pibd} and MMP \cite{mmp}, offer solutions to reduce redundancy, they only use simple survival follow-up labels to select discriminative features. However, these labels offer limited supervision, which is insufficient for such a complex survival prediction task influenced by multifaceted factors.

To address above challenges , we propose LLM-driven Knowledge-Enhanced Multimodal Model (KEMM), where expert reports and prognostic background knowledge (PBK) collaboratively enhance two highly redundant modalities to extract discriminative features and offer adequate prognostically focused knowledge on the complex survival prediction task. 
First, expert reports, written by pathologists on a case-by-case basis and refined by large language model (LLM), clearly provide valuable diagnostic information closely related to patients' outcomes. For example, pathology morphological subtypes and immunohistochemical markers that inform gene expression profiles, potentially indicate different survival outcomes \cite{sun2025pathgenm,Guo_HistGen_MICCAI2024}.
Second, prognostic background knowledge generated by LLM, provides additional prognostic contexts that are not included in expert reports, such as diverse manifestations across different survival risk levels reflected in both pathology and genomic data, which further enhance the survival prediction task. 
Together, these two complementary texts provide more prognostically focused and comprehensive knowledge than conventional survival labels. 
As a result, condensed and synergistic external knowledge is established to guide the learning of highly redundant data and improve survival prediction.

The contributions of this work can be summarized as follows: 
\begin{itemize}
    \item[$\bullet$] We propose LLM-driven Knowledge-Enhanced Multimodal Model for cancer survival prediction, \textit{i.e.}, KEMM, where expert knowledge and prognostic background knowledge enhance the learning of highly redundant pathology and genomic data with sufficient prognostically focused information for complex multimodal survival prediction.
    \item[$\bullet$] We design knowledge-enhanced cross-modal (KECM) attention modules, which enable external prognostic knowledge to guide redundant modalities modeling, thereby aligning heterogeneous modalities into a condensed space and extracting survival-related and discriminative features.
    \item[$\bullet$] Extensive experiments on five cancer datasets demonstrate results significantly superior to state-of-the-art (SOTA) methods.
\end{itemize}

\section{Method}

\begin{figure}[t]

    \centering
    \includegraphics[scale=0.34]{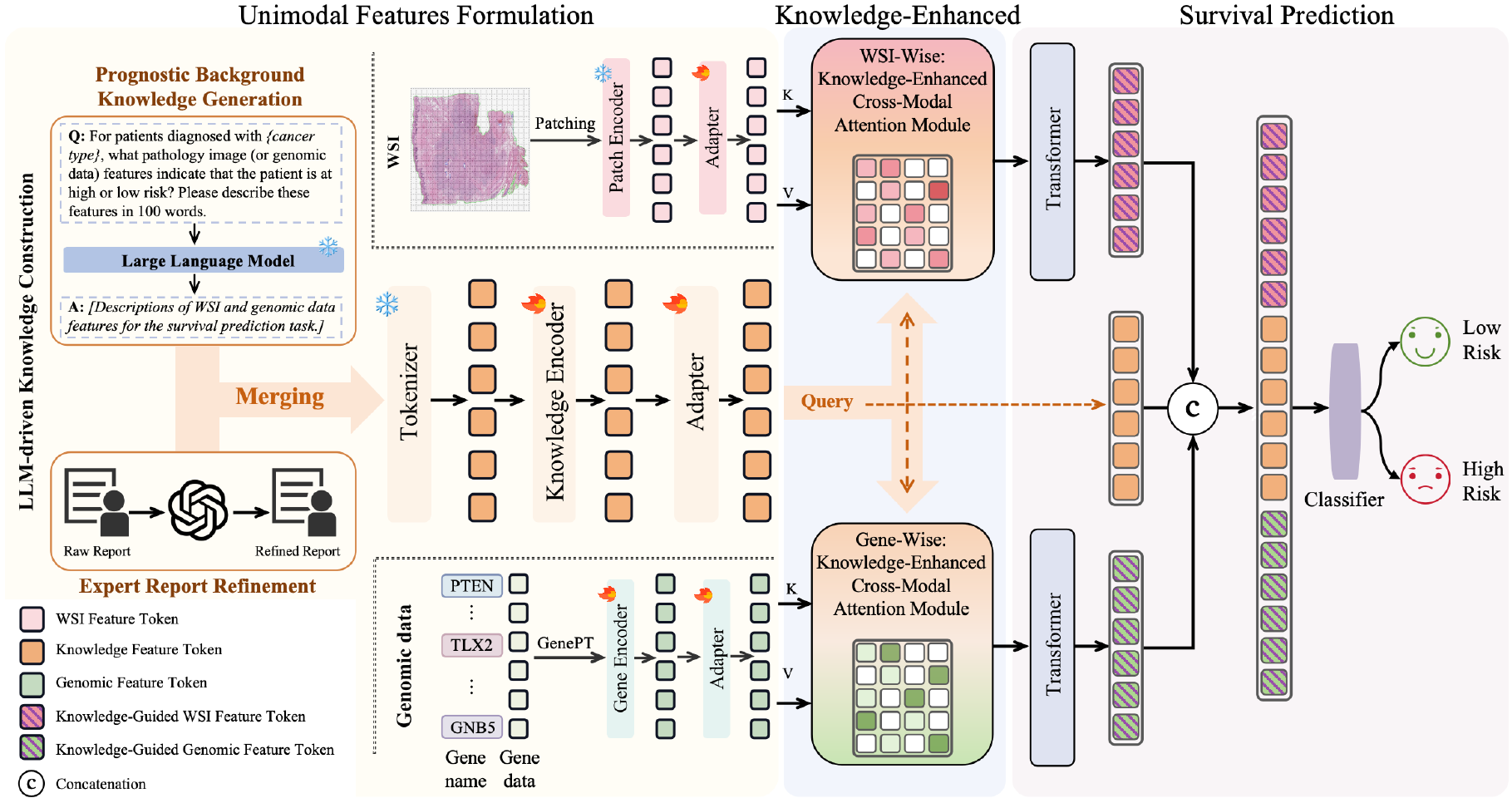}
    \caption{Overview of KEMM. Left: Prognostic Background Knowledge Generation, Expert Report Refinement, and each modality features formulation. Middle: Knowledge-Enhanced Cross-Modal Attention to enhance WSI and genomic features. Right: multimodal features aggregation for survival outcome prediction.}
    \label{network}

\end{figure}

This work aims to introduce expert and prognostic background knowledge as guidance for the WSIs and genomic data and enhance multimodal survival prediction.
The overall framework of KEMM is shown in Fig.\ref{network}. 
We first introduce the problem formulation in Sec.
\ref{problem formulation}, and the encoding process for unimodal features is in Sec. \ref{unimodal}. Finally, the knowledge-enhanced multimodal survival prediction method is in Sec. \ref{multimodal}.

\subsection{Problem Formulation}
\label{problem formulation}
For a multimodal input of KEMM, the data is organized as 4-tuples, $ X = (P , G , R , PBK )$ for one patient, where $P  $ denotes the pathology image, $G $ is the genomic data, $R$ denotes the expert report provided by the doctor, and $PBK$ is the prognostic background knowledge (PBK) generated by the LLM, respectively.
In survival prediction, the aim is to estimate the hazard function $h(t)  = h(T=t|T \geq t, X)\in [0,1]$, representing the probability of death at the time point $t$. Subsequently, the survival probability after the time point $t$ is calculated using the cumulative hazard function, $i.e.$, $s(t|X)=\prod_{j=1}^{t}(1-h(j) )$.

\subsection{Unimodal Features Formulation}
\label{unimodal}

\begin{flushleft}
\textbf{Pathology Features Encoding.} Due to the huge size of WSIs, following previous work \cite{shao2021transmil,li2021dual,ilse2018attention}, we formulate its learning process as a weakly supervised Multiple Instance Learning (MIL) task. For each WSI, denoted as $P $, we first crop a set of non-overlapping patches that are embedded into features by frozen pretrained encoder $E_P(\cdot)$, leading to a bag of patch features $F_{patches} \in \mathbb{R}^{n_p \times d_{patch}}$. $ n_p$ is the number of patches and $ d_{patch}$ is the feature dimension of each patch. Then, we use an adapter $f_P(\cdot)$ to project features into latent space. As a result, a WSI are represented as $F_{P} = f_P(E_P(P)) \in \mathbb{R}^{n_p \times d}$, where $d$ is the feature dimension after the adapter.
\end{flushleft}

\begin{flushleft}
\textbf{Genomic Features Encoding.} Genomic data $G = (g, v)$ is organized as a 2-tuple with the gene name ($g$) and its expression variable ($v$). 
Following previous work \cite{yang2022scbert}, we employ GenePT \cite{chen2024genept} to generate gene embeddings ($g_e \in \mathbb{R}^{n_G \times d_G}$) for each gene name ($g$) in the biological context, where $n_G$ is the number of genes and $d_G$ is embedding dimension after GenPT. 
Additionally, a gene value encoder $E_v(\cdot)$ also embeds gene expression variable ($v$) into space of $\mathbb{R}^{n_G \times d_G}$.
These two embeddings are then element-wise added, resulting in a genomic input.
Subsequently, a genomic encoder $E_G(\cdot)$ and adapter $f_G(\cdot)$ embed them into a latent space. Consequently, the genomic features are formulated as $F_{G }  = f_G(E_G(\text{GenePT}(g)+E_v(v)) \in \mathbb{R}^{n_G \times d}$.
\end{flushleft}

\begin{flushleft}
\textbf{LLM-driven Knowledge Generation.} In KEMM, we incorporate expert knowledge from pathologists reports and prognostic background knowledge (PBK) generated by LLM (\textit{e.g.}, GPT-4 \cite{gpt4}) to enhance the modeling of pathology and genomic data.
Given that raw reports from The Cancer Genome Atlas (TCGA) are dirty and have diverse structural templates, we refine them with LLM, \textit{i.e.}, GPT-4, to the ones with a uniform structure following the prompt in \cite{mstar}, yielding refined report ($R$). 
For the generation of PBK, we design survival-related question prompts for both pathology images and genomic data based on cancer type, as follows: 
\textit{For patients diagnosed with 
\{cancer type\}, what pathology image (or genomic data) features indicate that the patient is at high or low risk? Please describe these features in 100 words.} 
After passing prompts to LLM, concise and valuable descriptions of pathology images and genomic data features across different risk levels are generated. Then, these two pieces of texts are merged along the word dimension into $K  = (R , PBK ) \in \mathbb{R}^{n_K}$ to form a condensed and synergistic knowledge, where $n_K$ is the token number of the merged text.
\end{flushleft}

\begin{flushleft}
\textbf{Knowledge Features Encoding.} Knowledge $K  = (R , PBK )$ is embedded by a fine-tuning pretrained LLM encoder ($E_K(\cdot)$)~\cite{BioClinicalBERT} into $d_K$-dimension features for each word. A knowledge adapter, $f_K(\cdot)$, is then applied, producing the $d$-dimensional knowledge features $F_{K} = f_K(E_K(K )) \in \mathbb{R}^{n_K\times d}$.
\end{flushleft}

\subsection{Knowledge-Enhanced Multimodal Survival Prediction}
\label{multimodal}
This section introduces knowledge-enhanced multimodal survival prediction, specifically focusing on the knowledge-enhanced cross-modal attention module (KECM) for WSI and genomic features. As mentioned, the external knowledge including expert insights for each cancer patient and prognostic background knowledge on different cancer types, can guide the extraction of discriminative multimodal feature tokens, aligning them into a more condensed space.

\begin{flushleft}
\textbf{Knowledge-Enhanced Cross-Modal Attention Module (KECM).} Following the Transformer attention layer \cite{waswani2017attention}, KECM uses knowledge features $F_{K}$ as guidance to query WSI patch features $F_{P }$ and genomic features $F_{G }$, respectively. 
Applying this guidance process to both pathology and genomic features, we obtain the knowledge-enhanced pathology features and genomic features, which can be respectively represented as follows:
\begin{equation}
    F_{P\to K}  = \text{norm}\Big(\text{softmax}\Big(\frac{W_{q}^{P} F_{K} F_P^{\top} {W_{k}^{P}}^\top }{\sqrt{d}}\Big)\Big)W_v^{P}F_P,
\label{wsi-text}
\end{equation}
\begin{equation}
    F_{G\to K} = \text{norm}\Big(\text{softmax}\Big(\frac{W_{q}^{G} F_{K} F_G^{\top} {W_{k}^{G}}^\top }{\sqrt{d}}\Big)\Big)W_v^{G}F_G,
\label{gene-text}
\end{equation}
where $F_{P\to K} \in \mathbb{R}^{n_K \times d}$ and $F_{G\to K} \in \mathbb{R}^{n_K \times d}$ denote the knowledge-enhanced pathology and genomic feature, respectively. $W_q^{P}$, $W_k^{P}$, $W_v^{P} \in \mathbb{R}^{d \times d}$ are trainable weight matrices for query $F_K$ and key-value pair $(F_P, F_P)$.
Similarly, $W_q^{G}$, $W_k^{G}$, and $W_v^{G} \in \mathbb{R}^{d \times d}$ are used for the query $F_K$ and the key-value pair $(F_G, F_G)$.
Thus, two heterogeneous features are aligned into a compact space.
\end{flushleft}

\begin{flushleft}
\textbf{Multimodal Survival Prediction.} 
After KECM, a Transformer ($\mathcal{T}$) encoder will integrate all tokens after enhancement for pathology and genomic features, respectively.
All modalities features will be concatenated to final multimodal features $F_{final} = \mathcal{T}_P(F_{P\to K})\bowtie F_K\bowtie \mathcal{T}_G(F_{G\to K}) \in \mathbb{R}^{n_K \times 3d}$, where $\mathcal{T}_P, \mathcal{T}_G$ are transformers for knowledge-enhanced pathology and genomic features.
After concatenation, the mean pooling applies to obtain the final multimodal token for classification, \textit{i.e.}, $\rho: \mathbb{R}^{n_K \times 3d} \Rightarrow \mathbb{R}^{3d}$. After classifier, the hazard function $h(t)$ and survival probability $s(t|F_{final})$ will be calculated according to Sec. \ref{problem formulation}. Finally, following the previous work \cite{motcat,mcat}, negative log-likelihood (NLL) loss \cite{nllloss} is employed to optimize the framework:
\begin{equation}
\small
    \mathcal{L}_{surv}= -\sum_{i=1}^{N}\Big(c_i \log (s_{i}(t|F_{final}^i)) + (1-c_i)\log(s_i(t-1|F_{final}^i)) + (1-c_i)\log(h_i(t|F_{final}^i)) \Big),
\label{loss}
\end{equation}
where $N$ is the number of patients, $c_i \in \{0,1\}$ is censorship for the $i^{\text{th}}$ patient.
\end{flushleft}
\section{Experiment}

\begin{table}[htbp]
\small
\centering
\setlength{\tabcolsep}{5pt}
\renewcommand\arraystretch{1.2}
\caption{Results on 5 datasets. Best performances are in \textcolor{red}{red} and second-best are in \textcolor{blue}{blue} among all methods, while the best in each group are in \textbf{bold}. $P$ is WSI, $G$ is genomic data, $R$ is report, and $PBK$ is prognostic background knowledge.}
\label{allresults}
\arrayrulecolor{black}
\resizebox{12.2cm}{!}{
\begin{tabular}{cccc!{\color{black}\vrule}l!{\color{black}\vrule}ccccc!{\color{black}\vrule}c} 
\toprule[1.5pt]
\multicolumn{4}{c!{\color{black}\vrule}}{Data} & \multirow{2}{*}{\makecell[c]{ Method}} & {BRCA} & {LUAD} & {UCEC} & {LUSC} & {KIRC} & \multirow{2}{*}{AVG} \\ 
\arrayrulecolor{black}\cline{1-4}
\scriptsize{$P$} & \scriptsize{$G$} &\scriptsize{$R$} & \scriptsize{$PBK$} & & $(N = 1007)$ & $(N = 443)$ & $(N = 478)$ & $(N = 434) $ & $(N = 473)$ & \\ 
\hline\hline
\Checkmark & ~ & ~ & ~ &  MeanMIL & $0.5867\pm 0.031$ & $0.6013\pm 0.078$ & $0.6376\pm 0.063$ & $0.5928\pm 0.061$ & $0.6472\pm 0.029$ & $0.6131$ \\ 
\Checkmark & ~ & ~ & ~ &  MaxMIL & $0.5690\pm 0.044$ & $0.5796\pm 0.039$ & $0.5485\pm 0.039$ & $0.5554\pm 0.014$ & $0.6534\pm 0.013$ & $0.5812$ \\ 
\Checkmark & ~ & ~ & ~ &  TransMIL \cite{shao2021transmil} & $0.5882\pm 0.057$ & $\bm{0.6351\pm 0.035}$ & $0.6273\pm 0.035$ & $\bm{0.6081\pm 0.048}$ & $\bm{0.6632\pm 0.057}$ & $\bm{0.6244}$ \\ 
\Checkmark & ~ & ~ & ~ &  AttnMIL \cite{ilse2018attention} & $\bm{0.5976\pm 0.042}$ & $0.6042\pm 0.077$ & $0.6381\pm 0.063$ & $0.5905\pm 0.071$ & $0.6479\pm 0.017$ & $0.6157$ \\ 
\Checkmark & ~ & ~ & ~ &  DSMIL \cite{li2021dual} & $0.5874\pm 0.035$ & $0.5979\pm 0.078$ & $\bm{0.6384\pm 0.068}$ & $0.5909\pm 0.062$ & $0.6529\pm 0.029$ & $0.6135$ \\ 
\hline\hline
~ & \Checkmark & ~ & ~ &  MLP & $0.6934\pm 0.021$ & $0.6438\pm 0.072$ & $\bm{0.7245\pm 0.094}$ & $0.5547\pm 0.043$ & $\bm{0.7291\pm 0.031}$ & $0.6691$ \\ 
~ & \Checkmark & ~ & ~ &  SNN \cite{snn} & $\bm{0.6959\pm 0.031}$ & $\bm{0.6613\pm 0.060}$ & $0.7241\pm 0.076$ & $\bm{0.5572\pm 0.039}$ & $0.7255\pm 0.031$ & $\bm{0.6728}$ \\ 
~ & \Checkmark & ~ & ~ &  GenePTrans \cite{chen2024genept} & $0.6457\pm 0.044$ & $0.6247\pm 0.050$ & $0.7035\pm 0.056$ & $0.5490\pm 0.046$ & $0.6935\pm 0.019$ & $0.6433$ \\ 
\hline\hline
~ & ~ & \Checkmark & ~ &  BioBERT \cite{lee2020biobert} & $0.7043\pm 0.033$ & $0.6326\pm 0.037$ & $0.7003\pm 0.074$ & \textcolor{blue}{$\bm{0.6161\pm 0.006}$} & \textcolor{red}{$\bm{0.7778\pm 0.022}$} & $0.6862$ \\ 
~ & ~ & \Checkmark & ~ &  BioClinicalBERT \cite{BioClinicalBERT} & \textcolor{blue}{$\bm{0.7136\pm 0.024}$} & $0.6308\pm 0.040$ & $\bm{0.7318\pm 0.054}$ & $0.6019\pm 0.023$ & $0.7664\pm 0.045$ & \textcolor{blue}{$\bm{0.6889}$} \\ 
\hline\hline
\Checkmark & \Checkmark & ~ & ~ &  MCAT \cite{mcat} & $0.7013\pm 0.031$ & $0.6592\pm 0.065$ & $0.7253\pm 0.065$ & $0.5614\pm 0.025$ & $0.7248\pm 0.021$ & $0.6744$ \\ 
\Checkmark & \Checkmark & ~ & ~ &  MOTCat \cite{motcat} & $0.6858\pm 0.026$ & $0.6608\pm 0.057$ & $0.7260\pm 0.054$ & $0.5702\pm 0.019$ & $0.7378\pm 0.022$ & $0.6761$ \\ 
\Checkmark & \Checkmark & ~ & ~ &  CMTA \cite{cmta} & $\bm{0.7132\pm 0.012}$ & $0.6456\pm 0.055$ & \textcolor{blue}{$\bm{0.7386\pm 0.059}$} & $0.5749\pm 0.033$ & $\bm{0.7456\pm 0.01}$ & $\bm{0.6836}$ \\ 
\Checkmark & \Checkmark & ~ & ~ &  Porpoise \cite{porpoise} & $0.7077\pm 0.034$ & $0.6423\pm 0.065$ & $0.7229\pm 0.075$ & $0.5529\pm 0.042$ & $0.7389\pm 0.028$ & $0.6729$ \\ 
\Checkmark & \Checkmark & ~ & ~ &  SurvPath \cite{survpath} & $0.6744\pm 0.039$ & $0.6471\pm 0.041$ & $0.7041\pm 0.070$ & $\bm{0.6082\pm 0.025}$ & $0.7355\pm 0.030$ & $0.6739$ \\ 
\Checkmark & \Checkmark & ~ & ~ &  PIBD \cite{pibd} & $0.6741\pm 0.027$ & \textcolor{blue}{$\bm{0.6632\pm 0.033}$} & $0.7123\pm 0.082$ & $0.5882\pm 0.023$ & $0.7326\pm 0.047$ & $0.6741$ \\ 
\hline\hline
\Checkmark & \Checkmark & \Checkmark & \Checkmark &  Transformer (Cat) & $0.7098\pm 0.031$ & $0.6569\pm 0.062$ & $0.6831\pm 0.066$ & $0.6313\pm 0.042$ & $0.7452\pm 0.025$ & $0.6853$ \\ 
\Checkmark & \Checkmark & \Checkmark & \Checkmark &  Transformer (BP) & $0.6369\pm 0.047$ & $0.6133\pm 0.083$ & $0.6536\pm 0.078$ & $0.5963\pm 0.061$ & $0.7332\pm 0.035$ & $0.6467$ \\ 
\rowcolor[HTML]{FADADE}
\Checkmark & \Checkmark & \Checkmark & \Checkmark &  KEMM (Ours) & \textcolor{red}{$\bm{0.7250\pm 0.038}$} & \textcolor{red}{$\bm{0.6777\pm 0.010}$} & \textcolor{red}{$\bm{0.7532\pm 0.049}$} & \textcolor{red}{$\bm{0.6319\pm 0.036}$} & \textcolor{blue}{$\bm{0.7736\pm 0.009}$} & \textcolor{red}{$\bm{0.7123}$} \\
\bottomrule[1.5pt]
\end{tabular}}
\end{table}

\subsection{Experiment Settings}
\begin{flushleft}
\textbf{Datasets.} To evaluate the performance of KEMM, we conducted a series of experiments on five cancers using public datasets from TCGA \footnote{TCGA: https://portal.gdc.cancer.gov/} including WSIs, bulk-RNASeq data and expert reports for overall survival prediction, \textit{i.e.},  Breast Invasive Carcinoma (BRCA), Lung Adenocarcinoma (LUAD), Uterine Corpus Endometrial Carcinoma (UCEC), Lung Squamous Cell Carcinoma (LUSC) and Kidney Renal Clear Cell Carcinoma (KIRC). We used preprocessed genomic data of TCGA from cBioPortal\footnote{cBioPortal: https://www.cbioportal.org/} database. The number of cases for each cancer type is $N$ in Tab. \ref{allresults}.
\end{flushleft}
\begin{flushleft}
\textbf{Evaluation Metric.} Following the common setting in previous works \cite{mcat,motcat,cmta}, we perform 5-fold cross-validation on each cancer dataset. The concordance index (C-index) and its standard deviation (std) are reported.
\end{flushleft}
\begin{flushleft}
\textbf{Implementation.} For each WSI, we use ImageNet-pretrained \cite{deng2009imagenet} ResNet-50 \cite{he2016identity} as the patch feature encoder $E_P(\cdot)$, with a Multilayer Perceptron (MLP) adapter $f_P(\cdot)$. For genomic data, we employ GenePT \cite{chen2024genept} for gene names, and remove those genes that are not included in the dictionary of GenePT. Gene encoder $E_G(\cdot)$ is Nystr{\"o}mformer \cite{xiong2021nystromformer} followed by an MIL adapter $f_G(\cdot)$. For the generation of PBK, GPT-4 \cite{gpt4} is applied. BioClinicalBERT \cite{BioClinicalBERT} serves as the textual encoder $E_K(\cdot)$, and its adapter is implemented as a fully-connected layer. During training, we adopt Adam optimizer with an initial learning rate of $2\times 10^{-4}$ for the WSI adapter $f_P(\cdot)$ and  $2\times 10^{-5}$ for all other modules, along with weight decay of $1\times 10^{-5}$. All experiments are trained for 30 epochs.
\end{flushleft}

\begin{figure}[t]
    \centering
    \includegraphics[scale=0.36]{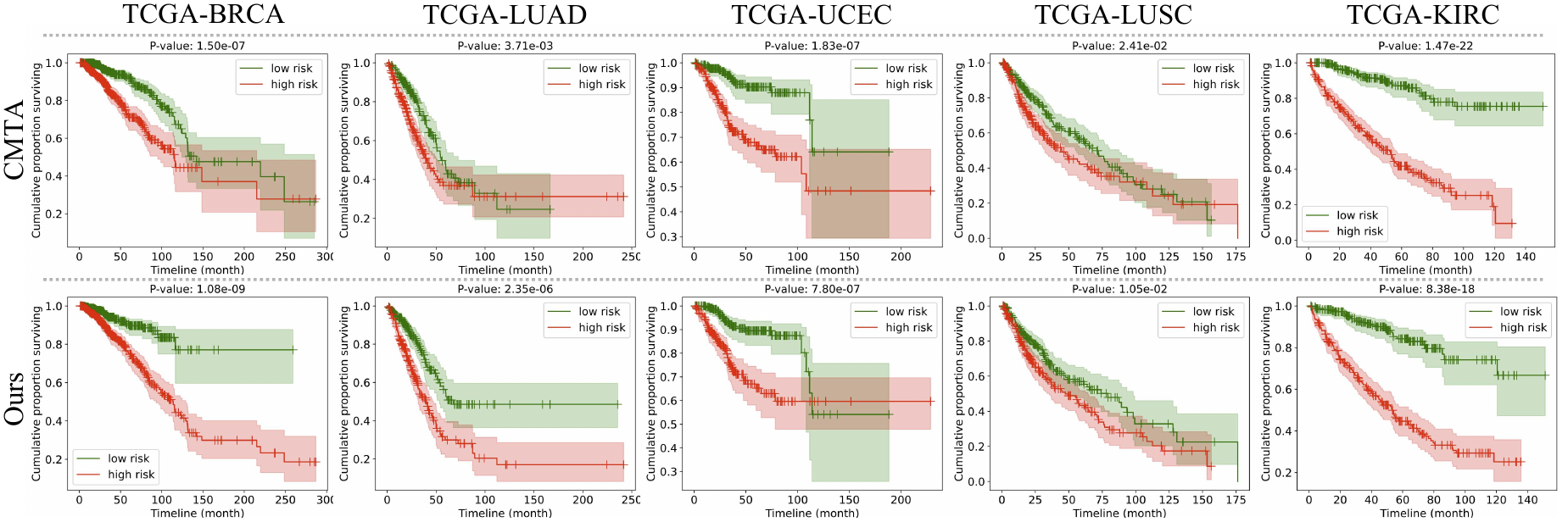}
    \caption{Kaplan-Meier curves of predicted high-risk (red) and low-risk (green) groups. A P-value $< 0.05$ indicates statistical significance between two groups.}
    \label{km_curve}
\end{figure}
\subsection{Comparison with SOTA Methods}
Results are shown in Tab. \ref{allresults}. For unimodal methods, we observe that the unimodal models for report superior both pathology and gene models, supporting the rationale for using textual knowledge to enhance features of other two modalities. Notably, unimodal methods for report outperform current multimodal fusion approaches in 4 out of 5 cancer datasets, indicating insufficient supervision in the previous multimodal survival methods. Our method surpasses unimodal models for reports in 4 datasets and achieves comparable results on KIRC.
For current SOTA multimodal methods, our results outperform all compared methods on 5 cancer datasets, demonstrating the effectiveness of incorporating external knowledge to enhance redundant modalities' learning by extracting survival-related and discriminative features.
Given that no compared methods that handle WSI, genomic data, and textual knowledge together, we design two baselines to evaluate the effectiveness of enhancement. For each input, we process the unimodal features as in KEMM and feed the WSI and genomic features into separate Transformer models. We then apply two common late fusion mechanisms, \textit{i.e.}, concatenation (Cat) and bilinear pooling (BP), namely Transformer (Cat) and Transformer (BP) in Tab. \ref{allresults}. These two results show that without knowledge enhancement, there is significant heterogeneity among different modalities, making fusion difficult, which indicates the necessity of an appropriate method for utilizing textual knowledge.
We also present Kaplan-Meier analysis and log-rank test for CMTA and KEMM in Fig. \ref{km_curve} to validate the statistical difference. KEMM separates two groups clearly and has a significant difference.

\subsection{Ablation Study}
\begin{flushleft}

\textbf{Contributions of different input data.} In Tab. \ref{different_input}, we assess the effect of different inputs, \textit{i.e.}, $(P,G,R,PBK)$. 
For the effect of $P$ and $G$, we compare our full version with methods that exclude either WSI or genomic data. As shown in Tab. \ref{different_input} (first part), removing any modality reduces performance, demonstrating the contribution of pathology and genes in survival prediction. 
Evaluation of $R$ and $PBK$ is shown in Tab. \ref{different_input} (second part). 
For methods excluding both report and PBK, compared with Transformer (Cat), the misalignment and heterogeneity between two modalities significantly impact the prediction accuracy. CMTA \cite{cmta} is a stronger pathology-gene baseline. KEMM can still outperform it, which indicates that CMTA struggles to align two redundant modalities, implying the importance of knowledge enhancement in reducing redundancy and aligning heterogeneous modalities.
When either $R$ or $PBK$ is removed, the full version performs the best in 4 of 5 data sets, confirming the synergy of expert and prognostic background knowledge.
\end{flushleft}
\begin{table}[t]
\small
\centering
\setlength{\tabcolsep}{5pt}
\renewcommand
\arraystretch{1.2}
\caption{Ablation on contributions of different input data. Best results are in \textcolor{red}{red}. Second-best results are in \textcolor{blue}{blue}.}
\arrayrulecolor{black}
\resizebox{12.2cm}{!}{
\begin{tabular}{cc!{\color{black}\vrule}cc!{\color{black}\vrule}l!{\color{black}\vrule}ccccc!{\color{black}\vrule}c} 
\toprule[1.5pt]
\multicolumn{4}{c!{\color{black}\vrule}}{Data} & \multirow{2}{*}{
\makecell[c]{Model}} & \multirow{2}{*}{BRCA} & \multirow{2}{*}{LUAD} & \multirow{2}{*}{UCEC} & \multirow{2}{*}{LUSC} & \multirow{2}{*}{KIRC} & \multirow{2}{*}{AVG} \\ 
\arrayrulecolor{black}\cline{1-4}
\scriptsize{$P$} & \multicolumn{2}{c}{\scriptsize{$G$\hspace{5mm}$R$}} & \scriptsize{$PBK$} &  & & & & & & \\
  \hline 
\multicolumn{11}{c}{\textit{Effect of WSI and gene modalities}} \\
\hline
\Checkmark &  & \Checkmark& \Checkmark& KEMM & $0.6952\pm0.054$ & \textcolor{blue}{$0.6489\pm0.025$} & $0.7252\pm0.058$ & \textcolor{blue}{$0.6242\pm0.031$} & $0.7618\pm0.013$ & 0.6911\\
 &  \Checkmark & \Checkmark& \Checkmark& KEMM  & $0.6993\pm0.042$ & $0.6715\pm0.026$ & $0.7291\pm0.066$ & $0.6004\pm0.025$ & $0.7668\pm0.032$ & 0.6934 \\
\hline
\multicolumn{11}{c}{\textit{Effect of expert and prognostic background knowledge}} \\
\hline
 \Checkmark & \Checkmark & & & Transformer (Cat)  & $0.6066 \pm 0.046$ & $0.6230 \pm 0.065$ & $0.6698\pm0.076$ &$0.5798\pm0.054$ &$0.7213\pm0.023$ & 0.6401 \\
\Checkmark & \Checkmark & & & CMTA \cite{cmta} & \textcolor{blue}{$0.7132\pm 0.012$} & ${0.6456\pm 0.055}$ & ${0.7386\pm 0.059}$ & $0.5749\pm 0.033$ & $0.7456\pm 0.011$ & 0.6836
 \\ 
 \Checkmark & \Checkmark & \Checkmark & & KEMM & {$0.7006\pm0.040$} & $0.6484\pm0.029$ & \textcolor{blue}{$0.7462\pm0.039$} & $0.6075\pm0.023$ & \textcolor{red}{$0.7824\pm0.045$} & \textcolor{blue}{0.6970} \\
  \Checkmark & \Checkmark & & \Checkmark & KEMM & $0.6494\pm0.045 $ & $0.6258\pm0.065$ & $0.7214\pm0.065$ & $0.5756\pm0.049$ & $0.7330\pm0.032$ & 0.6610 \\
\rowcolor[HTML]{FADADE}
\Checkmark & \Checkmark & \Checkmark & \Checkmark & KEMM (Ours) & \textcolor{red}{$0.7250\pm 0.038$}& \textcolor{red}{$0.6777\pm 0.010$} &\textcolor{red}{$0.7532\pm 0.049 $}& \textcolor{red}{$0.6319\pm 0.036$} & \textcolor{blue}{$0.7736\pm 0.009$} & \textcolor{red}{0.7123} \\
\bottomrule[1.5pt]
\end{tabular}}
\label{different_input}
\end{table}


\begin{table}[t]
\small
\centering
\setlength{\tabcolsep}{5pt}
\renewcommand
\arraystretch{1.2}
\caption{Results of different LLMs for PBK. Best results are in \textcolor{red}{red}. Second-best results are in \textcolor{blue}{blue}.}
\arrayrulecolor{black}
\resizebox{12.2cm}{!}{
\begin{tabular}{cccc!{\color{black}\vrule}l!{\color{black}\vrule}ccccc!{\color{black}\vrule}c} 
\toprule[1.5pt]
\multicolumn{4}{c!{\color{black}\vrule}}{Data} & \multirow{2}{*}{
\makecell[c]{ Method}} & \multirow{2}{*}{BRCA} & \multirow{2}{*}{LUAD} & \multirow{2}{*}{UCEC} & \multirow{2}{*}{LUSC} & \multirow{2}{*}{KIRC} & \multirow{2}{*}{AVG} \\ 
\arrayrulecolor{black}\cline{1-4}
\scriptsize{$P$} & \scriptsize{$G$} & \scriptsize{$R$} & \scriptsize{$PBK$} &  &  & & & & & \\ 
\hline\hline
\Checkmark & \Checkmark & ~ & ~ & CMTA \cite{cmta}& $0.7056\pm 0.024$ & $\textcolor{blue}{0.6677\pm 0.059}$ & $\textcolor{blue}{0.7359\pm 0.042}$ & $0.5763\pm 0.017$ & $0.7417\pm 0.014$ & ${0.6854}$ \\

\Checkmark & \Checkmark & \Checkmark & \Checkmark &  KEMM (HuatuoGPT) & \textcolor{red}{$0.7253\pm0.034$}& $0.6460\pm0.026$ & 
$0.7126\pm 0.032$ & $0.6039\pm 0.023$ & \textcolor{blue}{$0.7727\pm0.016$} & 0.6921 \\ 
\Checkmark & \Checkmark & \Checkmark & \Checkmark &  KEMM (DeepSeek-R1) & $0.7151\pm0.038$ & $0.6417\pm0.053 $ & 
$0.7233\pm 0.054$ & \textcolor{red}{$0.6342\pm 0.022$}& $0.7526\pm0.033$ & \textcolor{blue}{0.6934} 
\\
\rowcolor[HTML]{FADADE}
\Checkmark & \Checkmark & \Checkmark & \Checkmark &  KEMM (GPT-4, Ours) & \textcolor{blue}{$0.7250\pm 0.038$}& \textcolor{red}{$0.6777\pm 0.010$} & {\textcolor{red}{$0.7532\pm 0.049 $}}& {\textcolor{blue}{$0.6319\pm 0.036$}} & \textcolor{red}{$0.7736\pm 0.009$} & {\textcolor{red}{0.7123}}  \\
\bottomrule[1.5pt]
\end{tabular}}
\label{llm}
\end{table}

\begin{flushleft}
\textbf{Effect of different LLMs for PBK.} In Tab. \ref{llm}, we compare different large language models (LLMs) to generate PBK, \textit{i.e.}, GPT-4 \cite{gpt4}, HuatuoGPT \cite{huatuogpt-2023}, and DeepSeek-R1 \cite{deepseek}. KEMM can be compatible with different LLMs, all of which surpass CMTA \cite{cmta}. KEMM potentially benefits from stronger LLM.


\end{flushleft}
\section{Conclusion}
In this work, we propose a LLM-driven Knowledge-Enhanced Multimodal Model for survival prediction, named KEMM. External knowledge combines expert reports written by pathologists with prognostic background knowledge (PBK) generated by LLMs, providing succinct and valuable information for survival prediction. This knowledge serves as guidance to extract important features from highly redundant modalities, \textit{i.e.}, WSIs and genomic data. Besides, external knowledge can offer more sufficient and survival-related information than simple survival label on such a complex survival prediction task. Additionally, we introduce knowledge-enhanced cross-modal (KECM) attention module to enable the network to focus on discriminative and survival-relevant features. Extensive experiments on five cancer datasets demonstrate that our method achieves state-of-the-art performance.

\bibliographystyle{splncs04}
\bibliography{citation}
\end{document}